\documentclass[conference]{IEEEtran}
\IEEEoverridecommandlockouts
\usepackage{comment}
\usepackage{cite}
\usepackage{amsmath,amssymb,amsfonts}
\usepackage{algorithmic}
\usepackage{graphicx}
\usepackage{textcomp}
\usepackage{xcolor}
\usepackage{algorithm,algorithmic}
\usepackage{enumitem}
\usepackage{amsmath} 
\usepackage{lipsum} 
\usepackage{graphicx}
\usepackage[most]{tcolorbox}

\def\BibTeX{{\rm B\kern-.05em{\sc i\kern-.025em b}\kern-.08em
    T\kern-.1667em\lower.7ex\hbox{E}\kern-.125emX}}
\begin{document}

\title{A Hybrid Multi-Agent Prompting Approach for Simplifying Complex Sentences
}

\author{\IEEEauthorblockN{Pratibha Zunjare and Michael S. Hsiao}
\IEEEauthorblockA{\textit{Electrical and Computer Engineering, Virginia Tech} \\
\textit{Virginia Tech}, 
Blacksburg VA, USA \\
}

}

\maketitle

\begin{abstract}

This paper addresses the challenge of transforming complex sentences into sequences of logical, simplified sentences while preserving semantic and logical integrity with the help of Large Language Models. We propose a hybrid approach that combines advanced prompting with multi-agent architectures to enhance the sentence simplification process. Experimental results show that our approach was able to successfully simplify 70\% of the complex sentences written for video game design application. In comparison, a single-agent approach attained a 48\% success rate on the same task.
\end{abstract}


\section{\textbf{Introduction}}

\noindent Sentence simplification is a challenging task in computational linguistics.
 The simplification process aims to transform complex sentences into simpler structures while preserving the original meaning. Effective sentence simplification has significant applications across numerous domains like education, content accessibility for individuals with cognitive disabilities, automated content creation, robotics, coding, legal documents, etc.
  
Traditional approaches to sentence simplification have relied on rule-based systems, statistical methods, and more recently neural network architectures \cite{saggion2017automatic}. However, these methods often struggle with maintaining semantic equivalence while achieving appropriate simplification levels.

Complex sentences present significant challenges in action-oriented contexts, particularly when attempting to derive executable/actionable functionalities such as robotics, legal documents, and video games. These sentences frequently combine conditional logic (cause-effect constructs), nested dependencies, and multifaceted instructions within intricate syntactic structures. The crafting of unambiguous logic involves converting complex instructions written in natural language into precise, step-by-step actions. 
However, when sentences are complicated and interdependent on one another, advanced simplification and rewriting techniques are needed, which involve strategically breaking down complex sentences into simple, manageable steps while maintaining the underlying cause-and-effect relationships and procedural flow.
For example, consider the following sentence:

\textit{When a yellow alien touches a wall, it has 5 seconds before it explodes.}

While this sentence is comprehensible to human readers, it contains a temporal constraint of 5 seconds that demands the involvement of one or more intermediate states. This intermediate state helps in arbitrating when the alien touches the wall and when it explodes. Using the running example, one possible sequence of simplified sentences is as follows: \\ 

\textit{When a yellow alien touches a wall, the alien becomes sad for 5 seconds, and the alien becomes ready\_to\_explode for 6 seconds. When the alien is ready\_to\_explode and the alien is not sad, it explodes.} \\
 
The implicit delay (5 seconds) is transformed into explicit state changes with defined durations (\textit{sad for 5 seconds}, \textit{ready\_to\_explode for 6 seconds}). The resulting action (explosion) is tied to explicit state conditions rather than an implicit timer. 

While one might try to ask a Large Language Model (LLM) to perform the re-writing task, the vast variation of simplification strategies renders LLMs ineffective. Likewise, fine-tuning can also result in overfitting due to the limited training set. As a result, something new is needed to produce satisfactory results.

\begin{figure}[!htb]
  \centering
  \includegraphics[width=0.5\textwidth]{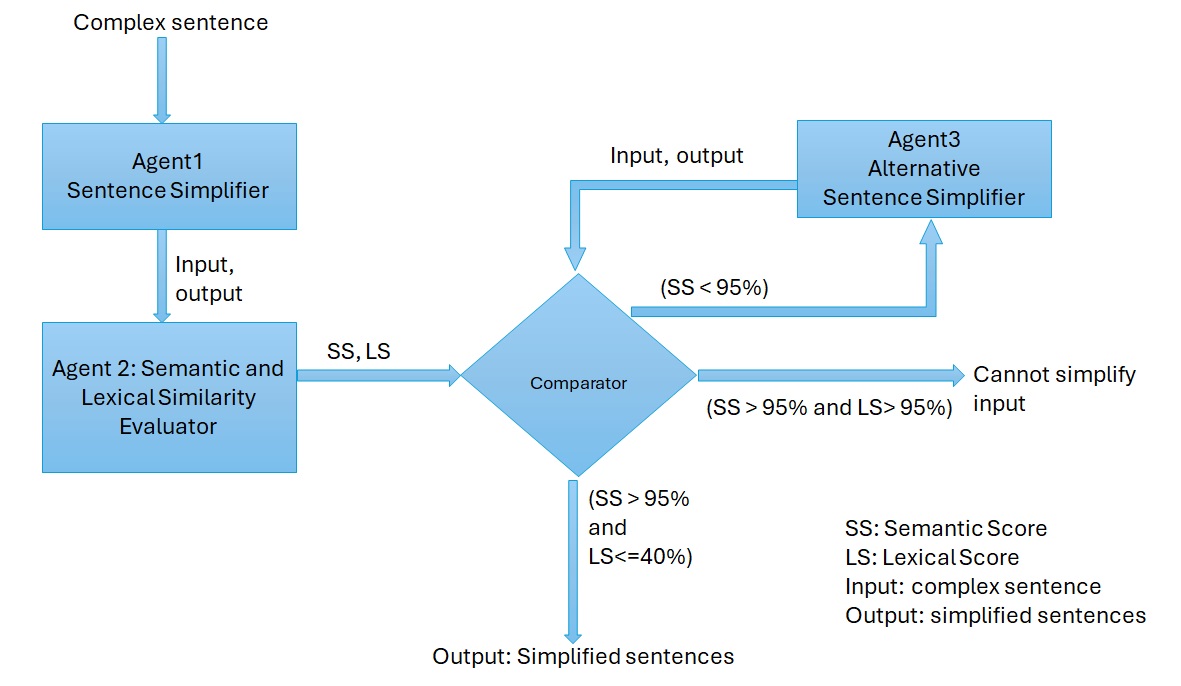} 
  \caption{Multi-Agent promptig workflow for sentence simplification}
  \label{fig:workflow}
\end{figure}

This paper introduces a novel approach, whose flow-chart is shown in Figure~\ref{fig:workflow}, that uses a multi-agent system with advanced prompt engineering to produce simplified sentences. We demonstrate how breaking down complex sentences into clear, actionable sentences enhances both understanding and execution. Our key contributions are: (1) a framework for optimizing prompts using multi-agent collaboration, (2) methods  \newpage \vspace*{0.05in} \noindent for translating complex instructions into executable steps, and (3) empirical evaluation of these techniques across a range of scenarios for video game design applications.


While LLMs may be effective in generating simple games, they often struggle with complex games with complex logic.
GameChangineer \cite{gameChangineer1} is an educational platform designed to help students develop fundamental logic and critical thinking skills. The proposed simplification process can serve as a learning tool to help students decompose complex scenarios into logical steps, a necessary precursor to learning to code. This platform is also used to evaluate the simplified sentences against the corresponding generated games.
The results show that the proposed simplification method was able to successfully decompose and simplify 70\% of the complex sentences for gaming applications. In contrast, a single-agent approach achieved a 48\% success rate on the same task.

The rest of the paper is organized as follows. Section II reviews the relevant background. Section III presents a comprehensive description of the proposed hybrid multi-agent prompting framework. Section IV analyzes the experimental results, comparing the performance of LLM-generated games using both simplified and complex sentences on the selected dataset. Section V concludes the paper and discusses directions for future work.

 \section{\textbf{Background}} 
 \subsection*{\textbf{Sentence Simplification }}
\noindent Modern approaches to sentence simplification are predominantly data-driven, utilizing parallel corpora of aligned complex and simplified sentences to learn effective simplification transformations, such as word substitution, sentence splitting, and reordering\cite{Manchego}. Despite progress, today's LLMs still struggle with semantic and lexical challenges in text simplification \cite{Feng2023simplification}. 
 \subsection*{\textbf{LLMs for Sentence Simplification}}
 \noindent LLMs have achieved remarkable capabilities in recent years on natural language processing tasks like text generation and summarization,  sentiment analysis, information retrieval and question answering, machine translation, etc., presenting researchers with diverse methodologies to fully leverage their potential.
 LLMs like \textsc{GPT-4o}, \textsc{GPT-4.1} \cite{OpenAI2024}, Gemini\cite{gemini2023}, Perplexity\cite{perplexity2025}, etc. can engage in fluent, contextual, and some reasonable conversations. As LLMs continue to evolve, researchers have developed numerous approaches to fully harness their potential, including prompt engineering, fine-tuning, and multi-agent systems \cite{Guo2024}. Despite these advances, certain linguistic challenges remain particularly difficult, especially those requiring nuanced understanding and transformation of language, including transforming complex sentences found in technical or operational domains—into actionable, logically coherent sentences \cite{Gooding}. Furthermore, large language models are prone to hallucination and often struggle to accurately break down sentences for precise game logic implementation \cite{agrawal2024knowledge}.

 \newpage \vspace*{0.1pt} \noindent
\subsection*{\textbf{Advanced Prompting Techniques}}
\noindent Prompt engineering refers to the systematic design of instructions to optimize an LLM's output for specific tasks. Effective prompts typically exhibit key characteristics including specificity, clarity, appropriate structure, and task decomposition \cite{Schulhoff2024}. 
Recent studies demonstrate that LLMs achieve superior performance when utilizing well-structured advanced prompting techniques, leveraging their inherent contextual reasoning capacities
  to produce more nuanced and reliable outputs compared to static instruction-based approaches\cite{Bsharat2024}. 
  Notable strategies include:
\begin{itemize}
\item
\textbf{Chain-of-Thought Prompting:} This technique encourages models to generate intermediate reasoning steps, 
improving their performance on tasks requiring multi-step inference and decomposition of complex instructions\cite{Wei2022}.
\item 
\textbf{In-Context Learning with Demonstrations:} By providing carefully curated examples within the prompt, 
LLMs can generalize to new domains and adapt to specific simplification tasks\cite{Dong2022}.
\item \textbf{Persona-Based Prompting:}
An advanced technique in prompt engineering that involves assigning a detailed role or persona to a LLM to guide its responses in terms of tone, style, reasoning and domain expertise. By specifying characteristics such as professional background, communication style, and knowledge boundaries, persona prompting enables the model to generate outputs that are more contextually relevant, consistent, and aligned with the intended use case\cite{Cheng2023}.
\item \textbf{Meta-Prompting and Dynamic Prompt Adaptation:} Recent work explores prompts that evolve dynamically based on model feedback or input complexity, further enhancing robustness in real-world scenarios\cite{CodaForno2023}.
\end{itemize}
As per the result from previous research, prompt design strongly influences the result of large language generative model \cite{Cheng2023}.
 These prompting methods can be used based on the type and complexity of the tasks.  There are various settings involved in the behavior of LLM for text generation like temperature and top-p. Temperature parameter controls the randomness to the model’s output whereas top-p controls the nucleus sampling which adds randomness to the model’s output \cite{Suzgun2024}. 
The design of the prompt and model’s setting plays a crucial role in the effectiveness of generated text.
\vspace{-2pt}
\subsection*{\textbf{Multi-Agent Architectures}}
\noindent While single-agent LLM offers remarkable fluency, they often struggle with both maintaining operational coherence and handling the hierarchical structure inherent in decomposition and simplification tasks. Multi-agent architectures address these limitations by distributing the simplification process across specialized agents, each responsible for distinct subtasks such as parsing, content extraction, logical validation, and refinement. Multi-agent architectures are categorized based on three key dimensions: the roles of LLMs as actors, the interaction strategies they employ, and the structural types of the systems\cite{Tran2025}.
\begin{itemize}
\newpage \vspace*{0.05in}
    \item \textbf{Role-Based:}  Agents are assigned focused roles (e.g., decomposition, evaluations, validation), mirroring human collaborative workflows and reducing cognitive overload on any single model component\cite{Tran2025}.
    \item \textbf{Hierarchical and Decentralized Coordination: } Architectures range from centralized controllers orchestrating agent interactions to decentralized, self-organizing agent networks that adapt to task complexity \cite{Wang2025}.
    \item \textbf{Iterative Cross-Agent Validation:} Agents iteratively review and refine each other's outputs, leveraging contrastive prompting and feedback mechanisms to enhance output quality \cite{Han2024}.
\end{itemize}

\section{\textbf{Methodology}}

\noindent We propose a hybrid approach that uses multiple LLMs to break down complex sentences written in natural language into simple, logical, and coherent sentences. We specifically target complex sentences written for video game design application and ensure that the simplified sentences are acceptable by the GameChangineer engine \cite{gameChangineer1}. However, the proposed flow is applicable to other application domains as well. 

\begin{algorithm}
\caption{Hybrid Multi-Agent Sentence Simplification Workflow}
\label{alg:hybrid-simplification}
\begin{algorithmic}[1]
    \renewcommand{\algorithmicrequire}{\textbf{Input:} Complex sentence}
    \renewcommand{\algorithmicensure}{\textbf{Output:} Simple sentences}

\STATE \textbf{Agent 1: Sentence Simplifier}
\STATE $S \leftarrow$ \texttt{LLM\_Breakdown}(Q)
\COMMENT{LLM generates break-down sentences}

\STATE \textbf{Agent 2: Semantic and Lexical score Evaluator}
\STATE \texttt{semantic\_score} $\leftarrow$ \texttt{Semantic\_Evaluator}(S)
\STATE \texttt{lexical\_score} $\leftarrow$ \texttt{Lexical\_Evaluator}(S)
\COMMENT{Returns the semantic and lexical scores between input and output sentences} \\
\COMMENT{//Comparator block starts}

\IF{$semantic\_score > 95$ and $lexical\_score <= 40$}\label{step:loop}
    \RETURN $S$ \\
\ENDIF
\IF{$semantic\_score > 95$ and $lexical\_score > 95$}
        \RETURN cannot convert
\ENDIF
\IF{$semantic\_score < 95$}
    \STATE \textbf{Agent 3: Alternative sentence Simplifier}
    \STATE $S' \leftarrow$ \texttt{LLM\_RevisedBreakdown}(Q)
    \STATE Call Agent 2
    \STATE Go to step \ref{step:loop} 

\ENDIF \COMMENT{//Comparator block ends}

\end{algorithmic} 
\end{algorithm}

\noindent Our algorithm is based on advanced multi-agent prompting techniques. This framework coordinates with three specialized agents for the following tasks: (1) the decomposition of sentences, (2) the semantic and lexical evaluation, and (3) the iterative revision, to ensure both contextual equivalence and linguistic quality of the resultant sentences with input sentence.
 The workflow is illustrated in Algorithm 1.

\subsection*{\textbf{Agent Descriptions}}
 \newpage \vspace*{0.05in}
 \noindent \textbf{Agent 1: Sentence Simplifier}\\
    The sentence simplification agent receives the user query, 
   
    which typically consists of complex sentences that encompass multiple steps, interdependent causes, and corresponding actions.
    The input query is incorporated into advanced, carefully designed prompts for \textsc{GPT-4o} LLM, 
    which then decomposes the complex sentence into a sequence of simpler, logically ordered instructions. The meta-instruction-based prompt\cite{Bsharat2024} handles generalized form of complex sentences to decompose into simple, logical cause-action sentences. The output sentences are evaluated based on semantic and syntactic similarity with the input sentence. 
    For example, consider the example shown below in \textit{Example: Reasoning in Prompt1}.
\vspace{-0.01pt}
\begin{tcolorbox}
 [colback=gray!5!white, colframe=black!50, title=Example 1: Reasoning in Prompt1,before skip=5pt,    
    after skip=5pt , breakable]
\textbf{Query:} \textit{When the rabbit is yellow, the fox that is touched by the rabbit will die.}
\vspace{0.5em}
\textbf{Let's think step by step:}
\begin{itemize}
  \item \textit{Rabbit is yellow.} \hfill (cause)
  \item \textit{Rabbit touches the fox.} \hfill (cause)
  \item \textit{Fox dies.} \hfill (action)
\end{itemize}
\vspace{0.5em}
\textbf{Answer:} \textit{When the rabbit is yellow and the rabbit touches the fox, the fox dies.}
\end{tcolorbox}

In this example, we illustrate how a step-by-step reasoning process facilitates the logical and sequential decomposition of a complex sentence based on our Agent 1. The prompt is crafted such that the input query is examined to answer the separate causes and actions. The specifics of the prompt are provided in the provided page\footnote{https://github.com/pratibha-vt/CS\_Simplification}. 
Such an approach is crucial for translating natural language game specifications into unambiguous, sequential rules that can be reliably interpreted and executed by a game engine.
\noindent Consider another example,
\begin{tcolorbox}[
  title=Example 2: Temporal and Referential Sentence Simplification,
 colback=gray!5!white, colframe=black!50, boxsep=2pt,
  top=1pt,
  bottom=1pt,
  fonttitle=\bfseries,
  breakable,
  enhanced]
\textbf{Original Query:} \\
\textit{When a fox sees the rabbit touch a carrot, it chases it until the rabbit moves.}

\textbf{Transformation Steps:}
\begin{itemize}
  \item Resolve pronouns (\textit{it} $\rightarrow$ fox or rabbit).
    \item Separate embedded actions: ``sees the rabbit touch'' $\rightarrow$ two actions: 
\end{itemize}
 
  \begin{itemize}
      \item \textit{The fox sees the rabbit}
      \item \textit{The rabbit touches a carrot}
  \end{itemize}
  \begin{itemize}

  \item Make temporal dependency explicit by splitting into distinct conditions.
\end{itemize}
\textbf{Simplified Output (Sentence Simplifier Agent):}
\begin{itemize}
  \item \textit{When a fox sees the rabbit and the rabbit touches a carrot, the fox chases the rabbit.}
  \item \textit{When the rabbit moves, the fox stops chasing.}
\end{itemize}
\end{tcolorbox}


 \newpage \vspace*{0.05in} \noindent
\noindent The example 2 shows that pronouns are resolved and temporal dependencies are made explicit by dividing the sentence into two: one describing the chase and the other describing when it stops.

To promote clarity and logical coherence, we systematically examine each sentence to identify primary causal factors, actions, and dependencies. Complex statements are subsequently decomposed into single-action steps articulated in straightforward language, deliberately minimizing the use of technical terminology and modifiers. Conditional relationships are made explicit, and terms like ``increase" or ``decrease" are standardized. The output follows a strict format, presenting both the original query and the simplified answer. 
    As per our dataset, this agent handles most of various conditional sentences based on cause-effect sentence structure. \\

\noindent \textbf{Agent 2: Semantic and Lexical Similarity Evaluator} \\
\textit{Semantic similarity evaluator} checks how closely the meanings of input complex sentence and the generated simplified sentences align, regardless of their structure. It gives a score of 100 if the meaning of the input (complex) and output (simplified) sentences are equivalent; lower score shows divergence in the context.

\textit{Lexical similarity score} is calculated based on the structure of sentences by ignoring the similar words. Lower lexical scores are preferred because they represent the different structures between the simplified sentences and the input sentence. We provide 3 examples below.

\begin{tcolorbox}[title=Agent 2 Example 1, colback=gray!5!white, colframe=black!50, breakable]
\textbf{Query:} \textit{When a fox sees the rabbit touch a carrot, it chases it until the rabbit moves.}

\textbf{Answer:}
\begin{itemize}
  \item \textit{When the fox sees the rabbit and rabbit touches a carrot, the fox turns silver for 0.1 second.}
  \item \textit{When the fox is silver and a rabbit touches a carrot, the fox chases the rabbit.}
  \item \textit{When rabbit moves, the fox stops.}
\end{itemize}

\textbf{Semantic Similarity:} 100\\
\textbf{Lexical Similarity:} 40
\end{tcolorbox}

\begin{tcolorbox}[title=Agent 2 Example 2, colback=gray!5!white, colframe=black!50, breakable]
\textbf{Query:} \textit{When the rabbit is red, the fox cannot eat the rabbit.}

\textbf{Answer:} \textit{When the rabbit is red, the fox cannot eat the rabbit.} \\
\textbf{Semantic Similarity:} 100

\textbf{Lexical Similarity:} 100 \\
This pair is considered a \textit{CannotConvert sentence} as both sentences are identical.

\end{tcolorbox}
\newpage
\vspace*{0.05in}
\begin{tcolorbox}[title=Agent 2 Example 3, colback=gray!5!white, colframe=black!50, breakable]
\textbf{Query:} \textit{When a rabbit is touched, score adds 1.}

\textbf{Answer:} \textit{When a rabbit is touched, the score increases by 1.}

\textbf{Semantic Similarity:} 90\\
\textbf{Lexical Similarity:} 70

\end{tcolorbox}

These three examples above illustrate that while semantic similarity remains high (the meaning is preserved), lexical similarity can vary significantly depending on the structural changes introduced during simplification. \\
\\
\noindent\textbf{Agent 3: Alternative Sentence Simplifier}

\noindent The Alternative Sentence Simplifier is a context-aware language model agent designed to address specialized sentence simplification scenarios involving temporal dynamics and mathematical conditional logic. This approach integrates hybrid prompting by combining directional stimulus prompting, meta-prompting\cite{Suzgun2024}, and chain-of-thought (CoT) \cite{Wei2022} techniques. Unlike generic simplification approaches, rule-guided transformations are used to resolve ambiguities in sentences where outcomes depend on time-sensitive parameters or arithmetic constraints. \\
For instance, in interactive gaming narratives, the system accurately simplifies statements like ``The character's movement speed increases if 3 minutes pass without taking damage" by applying domain-specific rules that map temporal triggers (e.g., elapsed time thresholds) and mathematical conditions (e.g., damage counters) to unambiguous causal relationships. This targeted approach ensures faithful preservation of operational logic while reducing syntactic complexity, which is critical for applications requiring precision in dynamic environments. 

The Agent 3 is invoked only if the Agent 2 reports a semantic similarity less than 95\%, as shown in Algorithm \ref{fig:workflow}.

The specific prompts for Agents 1 and 3 were designed through an iterative process, during which each prompt was refined and assessed to ensure the generation of intended  simplified sentences are suitable for downstream code generation in game development contexts.

This agent specializes in analyzing complex sentences that involve dynamic mathematical relationships, such as time sensitive speed adjustments (e.g., ``increase the speed by 20\% every 5 seconds") or logic combinations requiring precise numerical reasoning (e.g., ``reduce the size by 15\% if the rabbit collects three carrot within 10 seconds"). These structures demand explicit handling of temporal dependencies, rate calculations, and conditional arithmetic operations inherent in game mechanics.

\subsection*{\textbf{Multi-Agent vs. Single-Agent Approach }}

\noindent While one might think that Agents 1 and 3 could be combined into a single agent, this might not work well because single  \newpage \vspace*{0.05in} \noindent agent models are constrained by the monolithic nature of single-prompt instruction handling. Thus, they often struggle to reconcile multiple objectives within a unified representation. As the complexity of simplification increases, these models tend to exhibit degraded performance and diminished interpretability \cite{Han2024}. The observed limitations of the single-agent formulation underscore the efficacy and practical benefits of adopting a multi-agent approach for complex sentence simplification.

In contrast, multi-agent approaches offer methodological advantages over single-agent systems, particularly in terms of modularity, adaptability, and scalability. By partitioning responsibilities across multiple specialized agents, the architecture supports fine-grained control over distinct subtasks, enabling more interpretable, maintainable, and extensible systems \cite{Wang2025}. This structural decomposition not only improves the performance but also enhances the system’s capacity to generalize across a broader spectrum of sentence types. 

\section{\textbf{Experimental Results}}
\subsection*{\textbf{Dataset}}
\noindent The dataset for sentence simplification in video gaming was meticulously curated by human annotators. A set of 100 complex sentences containing a range of linguistic and semantic features relevant to game logic and player interaction is used. Each sentence was evaluated for degree of complexity, presence of cause-effect relationships, descriptions of speed or size changes, and the occurrence of multiple actions within a single sentence. The dataset comprises a diverse distribution of sentence types, with conditional sentences forming the majority (50\%), followed by sequential (25\%), miscellaneous (15\%), and a notable subset labeled as \textit{‘Cannot Convert’} (10\%). Each category was defined: (1) conditional sentences specify actions contingent on in-game conditions (e.g., ``When the rabbit touches a gem, it gets more speed."); (2) sequential sentences mandate ordered execution of actions; (3) miscellaneous sentences encompass uncategorized patterns not fitting other classes. The \textit{‘Cannot Convert’} subset represents sentences where simplification risks critical information loss  or logical incoherence, reflecting inherent limitations in preserving game logic during rephrasing.  The curation process prioritized the identification and annotation of cause-and-action structures, as these are pivotal for downstream code generation tasks in gaming environments. For all of the experiments, \textsc{GPT-4o} was used.
\vspace{-0.1em}
\subsection*{\textbf{Results on dataset}}
\noindent The proposed multi-agent system achieved a successful simplification rate of 70\%, with these sentences exhibiting a semantic similarity greater than 95\% and a lexical similarity less than or equal to 40\%, indicating that the simplified sentences retained the original meaning while employing substantially different wording. Additionally, 10\% of the dataset was classified as Cannot Convert, reflecting cases where simplification would compromise essential information or logical structure. The  \newpage \vspace*{0.05in} \noindent \textbf{70\%} of sentences successfully simplified by the system underwent further human evaluation, confirming their acceptability for integration into the GameChangineer platform and ensuring alignment with the requirements of executable game logic.

The multi-agent prompting system fails to simplify 30\% of the sentences, particularly when handling abstract actions or ambiguous roles. For example, with \textit{``The cobras try to make sure the fox does not touch the bunny"} the system struggles to interpret ``try" and translate it into a concrete game action. Similarly, sentences like \textit{``It is not possible to win without eating all the carrots"} lack a clearly defined main player, leading to further simplification failures. 

We further evaluated system performance by disabling Agent 3 and relying solely on Agent 1 for simplification tasks. A single-agent approach achieved only a 48\% success rate on the same task.  Agent 1 struggled with sentences requiring analytical computations, such as \textit{``when the rabbit touches a diamond, its speed increases by 20\% every 10 seconds."} and was similarly unable to effectively simplify sentences involving positional and temporal reasoning, such as, \textit{``The rabbit starts at the bottom and moves up four times and drops a carrot every time it moves."} In contrast, Agent 3 effectively managed sentences with mathematical, spatial, and temporal complexities, underscoring the benefits of a multi-agent approach for simplifying intricate instructions.
\subsection*{\textbf{Game A: With Complex Sentences}}
\noindent The game below consists of complex sentences, such as those explained earlier in the paper. For generating code for the given game description, logical clarity is essential to derive the correct behavior of all the characters.

\begin{tcolorbox}[title=Game A (Complex Sentences), colback=gray!5!white, colframe=black!50, breakable]
\begin{itemize}[left=0pt]
    \item A fox and a rabbit wander in a field where 10 carrots and a diamond are scattered.
    \item Arrow keys control the rabbit.
    \item When a fox sees a rabbit but not a diamond, it chases the rabbit.

 \item When a fox is near, the speed of the rabbit increases  by 0.1.
    \item When a fox sees the rabbit move, it chases it until the rabbit touches a diamond.
    \item When the rabbit touches a diamond, it becomes yellow.
    \item When the rabbit is yellow, it can eat the fox it touches and you win.
    \item When a fox touches a diamond, it has 5 seconds before it explodes.
  
     \item When the fox explodes, the game is over.
    \end{itemize}
\end{tcolorbox}

\subsection*{\textbf{Game B: Simplified Version of Game A}}
\noindent The following version is derived using our multi-agent sentence simplification system. Sentences are made lexically simpler and structured for unambiguous logic.
\newpage
\vspace*{0.05in}
\begin{tcolorbox}[title=Game B (Simplified Sentences), colback=gray!5!white, colframe=black!50, breakable]
\begin{itemize}[left=0pt]
    \item There are 10 carrots, a diamond, a fox, and a rabbit.
    \item Carrots are scattered randomly.
    \item Fox and rabbit wander.
    \item When the left arrow is pressed, the rabbit moves left.
    \item When the right arrow is pressed, the rabbit moves right.
    \item When the up arrow is pressed, the rabbit moves up.
    \item When the down arrow is pressed, the rabbit moves down.
    \item When a fox sees a rabbit and does not see a diamond, the fox chases the rabbit.
    \item When a fox is near, the rabbit’s speed increases by 0.1.
    \item When a fox sees the rabbit and the rabbit moves, the fox chases the rabbit.
    \item When the rabbit touches a diamond, the fox stops chasing.
    \item When the rabbit is yellow and touches the fox, the rabbit eats the fox.
    \item When the rabbit eats the fox, you win.
    \item When a fox touches a diamond, it becomes sad for 5 seconds and ready\_to\_explode for 6 seconds.
    \item When the fox is ready\_to\_explode and not sad, it explodes.
    \item When the fox explodes, the game is over.
\end{itemize}
\end{tcolorbox}

\noindent
Game B statements are simplified for code generation with explicit state transitions and unambiguous lexical structure. The revised rules introduce phased behavioral states (e.g., \textit{sad}, \textit{ready\_to\_explode}) with temporally bounded conditions, creating a finite state machine structure that directly maps to programmable logic. 

Unlike Game A’s ambiguous temporal chain (e.g., ``5 seconds before explosion”), Game B uses clear state labels and logical triggers (e.g., \textit{ready\_to\_explode AND not sad}) to avoid race conditions. Explicit key mappings (left/right/up/down) also resolve vagueness, enabling direct keycode-to-method binding in implementation.

\subsection*{\textbf{Human Evaluation on Code generated for Game Description}}
\noindent Finally, we manually crafted more than ten game descriptions, each containing a subset of complex sentences from the data set alongside their simplified counterparts, to assess the impact of sentence complexity on the generation of code using LLMs. When generating game code from complex sentences, LLMs frequently missed constraints and misinterpreted instructions, resulting in games that ran longer but were inaccurate. Simplified sentences more effectively conveyed constraints, reducing ambiguity and improving code accuracy, yet LLMs can still struggle to consistently fulfill user intentions in both scenarios. Our evaluation indicates that while complex sentences  \newpage \vspace*{0.05in} \noindent increases ambiguity and reduces accuracy, the simplified sentences increases the number of constraints. Both complex sentences and increased number of constraints pose challenges to general-purpose LLMs.

More specifically, we find that LLMs struggle to generate precise code from games described in natural language that contain complex sentences as the LLMs need to rely on implicit assumptions in complex sentences and may not fully capture the intended logic. Conversely, when a game is rewritten with simplified sentences that offer greater number of constraints that capture both precision and explicit state conditions, the LLMs can overlook these critical conditions if the description contains too many constraints. 

Although general-purpose LLMs struggle generating these complex games, domain-specific engines such as the
GameChangineer \cite{gameChangineer2} platform not only generated more accurate games from simplified sentences, but also provided detailed debug messages and guidelines when it failed to understand a sentence. Such a feature can enable users to revise and correct errors in the game description. In contrast, general LLMs did not offer any error messages or feedback when encountering problematic sentences, making it difficult for users to identify and resolve issues in the generated code.

\section{\textbf{Conclusions and Future Work}}

\noindent Our approach leverages a hybrid multi-agent framework, wherein distinct LLM-based agents are assigned specialized roles such as sentence simplifier, lexical and semantic similarity evaluator, and alternative sentence simplifier. Two different sentence simplifying agents enable the system to address a wide variety of sentence types and linguistic phenomena, which are often challenging for single-agent models. By allowing agents to operate in parallel to enhance scalability and efficiency, as well as the overall quality of the simplification output. 
Empirical results show that our hybrid multi-agent  model achieves consistently stronger performance than single-agent baselines in both automatic and human evaluations for sentence simplification. 

A key challenge of sentence simplification is in simplifying sentences with abstract actions or ambiguous roles, which can often lead to failures in accurately conveying the game logic. Additionally, our current prompt design is closely tied to the dataset, which can lead to overfitting. Future work could address these by incorporating multiple agents to better handle more diverse sentence structures.

     
\end{document}